\documentclass{article}

\usepackage[final]{corl_2020} % Uncomment for the camera-ready ``final'' version

\usepackage{hyperref}
\usepackage[square,numbers]{natbib}
\usepackage{bm}
\usepackage{caption} 
\usepackage{subcaption}
\usepackage{amsmath}
\usepackage{graphicx}
\usepackage{makecell}
\usepackage{algorithm}% http://ctan.org/pkg/algorithm
\usepackage{algpseudocode}
\usepackage{wrapfig}
\usepackage[nolist,nohyperlinks]{acronym}
\usepackage{units}
\usepackage{mathtools}
\usepackage{multirow}
\usepackage{microtype}

\makeatletter
\newcommand{\printfnsymbol}[1]{%
  \textsuperscript{\@fnsymbol{#1}}%
}
\makeatother

\newcommand{\etal}{\textit{et al}. }
\DeclareMathOperator*{\argmax}{arg\,max}

\title{Learning Trajectories for Visual-Inertial System Calibration via Model-based Heuristic Deep Reinforcement Learning}

\author{
  Le Chen\thanks{equal contribution} \\
  ETH Zurich\\
  \texttt{lechen@ethz.ch} \\

  \And
  Yunke Ao\printfnsymbol{1} \\
  ETH Zurich\\
  \texttt{yunkao@ethz.ch} \\
   
  \And
  Florian Tschopp \\
  ETH Zurich\\
  \texttt{ftschopp@ethz.ch} \\
   
  \And
  Andrei Cramariuc \\
  ETH Zurich\\
  \texttt{crandrei@ethz.ch} \\
   
  \And
  Michel Breyer \\
  ETH Zurich\\
  \texttt{mbreyer@ethz.ch} \\
   
  \And
  Jen Jen Chung \\
  ETH Zurich\\
  \texttt{chungj@ethz.ch} \\
   
  \And
  Roland Siegwart \\
  ETH Zurich\\
  \texttt{rsiegwart@ethz.ch} \\
   
  \And
  Cesar Cadena \\
  ETH Zurich\\
  \texttt{cesarc@ethz.ch} \\

}

\begin{acronym}
\acro{mpc}[MPC]{model predictive control}
\acro{rl}[RL]{reinforcement learning}
\acro{pso}[PSO]{particle swarm optimization}
\acro{mdp}[MDP]{Markov decision process}
\acro{pomdp}[POMDP]{partially observed Markov decision process}
\acro{vi}[VI]{visual-inertial}
\acro{imu}[IMU]{inertial measurement unit}
\acro{fov}[FoV]{field of view}
\acro{sgd}[SGD]{stochastic gradient decent}

\end{acronym}

\begin{document}
\maketitle

%===============================================================================

\begin{abstract}
Visual-inertial systems rely on precise calibrations of both camera intrinsics and inter-sensor extrinsics, which typically require manually performing complex motions in front of a calibration target.
In this work we present a novel approach to obtain favorable trajectories for visual-inertial system calibration, using model-based deep reinforcement learning.
Our key contribution is to model the calibration process as a Markov decision process and then use model-based deep reinforcement learning with particle swarm optimization to establish a sequence of calibration trajectories to be performed by a robot arm.
Our experiments show that while maintaining similar or shorter path lengths, the trajectories generated by our learned policy result in lower calibration errors compared to random or handcrafted trajectories. \footnote[1]{The code is publicly available \url{https://github.com/ethz-asl/Learn-to-Calibrate}}
\end{abstract}

\keywords{Visual-Inertial, Calibration, Model-based Deep Reinforcement Learning, Markov Decision Process, Particle Swarm Optimization} 

%===============================================================================

\section{Introduction}

In recent years \ac{vi} sensors, which consist of one or more cameras and an \ac{imu}, have become increasingly popular for robust high frequency motion estimation~\citep{corke2007introduction,8633393,leutenegger2015keyframe,bloesch2017iterated,8421746}.
Before use, VI sensors need to be calibrated, which implies obtaining the parameters for the camera intrinsics, the camera-IMU extrinsics, and the time offset between the different sensors~\citep{tschopp2020versavis,nikolic2014synchronized}.
The performance of VI systems is highly dependent on the quality and accuracy of the calculated calibration parameters~\citep{kelly2011visual}.
Precise calibrations are usually obtained offline in controlled environments, following sophisticated motion routines ensuring observability~\citep{furgale2013unified,furgale2012continuous}.
This makes the entire process non-trivial for an inexperienced operator~\citep{kelly2011visual,nobre2019learning}.
Additionally, the motion primitives that most effectively render the best calibration results are unknown. 
So instead of performing this task by hand or with a manually programmed operator, we propose the use of model-based deep \ac{rl} to learn the best motion primitives and perform them on a robotic arm.

Previous works in automatic calibration use trajectory optimization and reinforcement learning to address this problem.
For the class of optimization methods that maximize the observability Gramian of the trajectories on the calibration parameters~\citep{preiss2017trajectory,hausman2017observability}, it remains challenging to model and include other practical optimization objectives, such as maximizing target point coverage for camera calibration and trajectory length minimization for efficiency.
A learning-based approach was proposed by Nobre \etal~\citep{nobre2019learning}, where a set of trajectories were pre-defined empirically and Q-learning was applied to choose a sequence of those trajectories that renders sufficient observability of the calibration problem.
A disadvantage of this approach is that it remains restricted to the collection of pre-defined trajectories and therefore the possible range of movements is not fully explored. 
We conclude that there are two aspects that could be further studied based on existing previous works:
\begin{itemize}
    \item \textbf{Multiple objectives:} It is not obvious whether the trajectory that renders sufficient observability of states also provides the best calibration. Besides, other empirical requirements such as path length and camera coverage could also be included in the optimized objectives.
    \item \textbf{Learn the trajectory:} A more general learning problem could be posed, where the predefined trajectories and their selection are learned jointly in a single optimization problem.
\end{itemize}

We aim to address both points by designing a \ac{rl}-based method to obtain the best sequence of \ac{vi} calibration trajectories that fulfill multiple objectives, including both observability and practical requirements.
The approach will be applied separately to calibrate both camera intrinsics and camera-IMU extrinsics, as these two steps have very different motion requirements.
The trajectories are executed by a robot arm during training in a realistic simulation for both dynamics and photorealism, thus guaranteeing the feasibility of execution by a real robot arm.
Additionally, we simulate different \ac{vi} sensor configurations to model the variability that also exists in the real world.

% To solve our problem, we model calibration as a partially observable Markov decision process (POMDP) and use RL to establish the sequence of motion trajectories which optimizes sensor calibration. Compared with other solutions, the action of our model has fewer constrains, making it possible to learn a more general policy for calibration. In addition, we not only consider different information-theoretic metrics of the trajectories but also take camera coverage and path length into account in the reward of our RL model. For the RL algorithms, Deep Deterministic Policy Gradient (DDPG)~\citep{lillicrap2015continuous} and Twin Delayed DDPG (TD3)~\citep{fujimoto2018addressing} are applied.  Different from Q-learning that can only learn policys in discrete action space, DDPG and TD3 are capable to learn a complex policy for continuous action spaces such as the trajectory parameter space. 

To solve the proposed problem, we model calibration as a \ac{mdp} and use model-based \ac{rl}~\citep{nagabandi2018neural} to establish the sequence of motion trajectories that optimizes sensor calibration accuracy.
Compared with other solutions the action space of our model has fewer constraints, making it possible to learn a more general policy for calibration.
In addition, we do not only consider different information-theoretic metrics of the trajectories, but also take camera coverage and path length into account, in the reward of our \ac{mdp} model.
For the learning algorithm, we propose a sample efficient model-based \ac{rl} method using the \ac{pso}~\citep{kennedy1995particle} algorithm to search for the optimal open-loop control sequence.
We adapt the \ac{pso} algorithm to \ac{rl} by utilizing gradient information and memorizing previous optimization results for initialization.
This method addresses difficulties in searching for action sequences in a high dimensional space and the high time cost of performing a calibration at each step. 

The main contributions of this work are as follows:
\begin{itemize}
\setlength\itemsep{0em}
\item Our approach models the entire calibration process as a \ac{mdp} and to the best of our knowledge, we are the first to solve it using model-based deep reinforcement learning.

\item Our proposed model-based heuristic \ac{rl} algorithm with adapted \ac{pso} satisfies different practical requirements such as the capability of solving problems with high dimensional action space with high sample efficiency. 
\item The evaluation shows that the learned trajectories deliver more accurate calibrations compared to handcrafted or random ones. We enable easy transferability to real scenarios by also simulating the robot arm together with different sensor configurations.
\end{itemize}

% \begin{Figure}[t]
%   \centering
%   % include first image
%   \includegraphics[width=1\linewidth]{framework.png}  
%   \caption{Overview of our calibration framework}
%   \label{fig:framework}
% \end{Figure}

% The overall structure of our paper is as follows: In Section 2, we review several related works and their limitations. In Section 3, we detail our method from problem formulation to algorithm design. In Section 4, we provide implementation details of our framework. Experimental results are shown in section 5. Finally, in Section 6, we summarize our contribution and discuss the drawbacks and possible future research directions.

%===============================================================================

\section{Related Work}
\label{sec:rel-work}

The most popular method for camera calibration during the last decades is to use a known calibration pattern and apply nonlinear regression to obtain the parameters.
This method has been successfully used both for camera intrinsic~\citep{sturm1999plane} and extrinsic calibration~\citep{zhang2004extrinsic}.
For VI sensor calibration the most reliable and precise approaches are also based on the use of a calibration board.
A method presented in~\citep{mirzaei2008kalman} applies an Extended Kalman Filter to estimate the relative poses between sensors jointly.
A parametric method proposed in~\citep{furgale2013unified,furgale2012continuous} represents the pose and bias trajectories using B-splines and introduces a batch estimator in continuous-time.
While those methods are relatively efficient, they still require expert knowledge to obtain the required sensor data for good accuracy.
In addition, the optimal calibration movements remain unknown, especially when aiming to increase time efficiency and calibration accuracy.
The problem of designing the best motion primitive was first addressed by methods based on trajectory optimization to find the trajectory that renders the highest observability of calibration parameters.

A method proposed in~\citep{hausman2017observability} optimizes the expanded empirical local observability Gramian of unknown parameters based on the measurement model, to solve for the best state and input trajectories.
Preiss \etal~\citep{preiss2017trajectory} further extended the method to be obstacle-free and more balanced among multiple objectives.
Both of the approaches only optimize observability by using the observability Gramian of the trajectories, where other empirical and general requirements are difficult to model and cannot be easily included.
Our framework combines different evaluation metrics of information gain for variables in the reward design and learns to obtain the highest reward using model-based \ac{rl}.

Another class of methods applies \ac{rl} to get the best sequence of trajectories selected from a library of pre-designed trajectories.
Nobre \etal~\citep{nobre2019learning} modeled the calibration process as an \ac{mdp}, where the states are estimated parameters and the actions are choices of trajectories from the library at each step.
The \ac{mdp} planning problem is solved using Q-learning.
However, the method only yields suggestions on predefined motions to choose from, rather than exploring new possible motion primitives.
In contrast, our method includes the trajectory parameters in the action design and solves the entire calibration problem in an end-to-end fashion.

For such complex sequential decision-making problems, recent works have shown that deep \ac{rl} algorithms are capable of learning policies that render high performance~\citep{mnih2015human,lillicrap2015continuous,schulman2017proximal,haarnoja2018soft,schulman2015trust}.
Model-based and model-free deep \ac{rl} are two classes of deep \ac{rl} algorithms.
Model-free algorithms are capable of learning a wide range of sequential decision problems, but they require a large number of samples to achieve good performance~\citep{schulman2015trust,feinberg2018model,buckman2018sample}. 
Nagabandi \etal~\citep{nagabandi2018neural} combined neural network model learning with sample-based \ac{mpc} to improve sample efficiency, and the policy is further fine-tuned with model-free algorithms.
Because of the high time cost to perform a calibration at each training step, model-based algorithms are suitable to reduce the number of required episodes to learn a good action sequence.

However, sample-based \ac{mpc} performs relatively poorly in high dimensional action space, therefore we substitute random sampling in~\citep{nagabandi2018neural} with a metaheuristic algorithm.
\ac{pso}~\citep{kennedy1995particle} is such a metaheuristic algorithm that has been widely used in the last two decades due to its good performance in complex high-dimensional problems, which cannot be solved using traditional deterministic algorithms.
Hein \etal~\citep{hein2016reinforcement} reformulated \ac{rl} problems as optimization tasks and applied \ac{pso} to search for optimal solutions.
In this paper, we combine \ac{mpc} with a modified \ac{pso} in a model-based framework to search for optimal action sequences. 

%===============================================================================

\section{Method}
\label{sec:approach}

\begin{figure}[t]
  \centering
  \setlength{\belowcaptionskip}{-0.5cm}
  % include first image
  \includegraphics[width=0.8\linewidth]{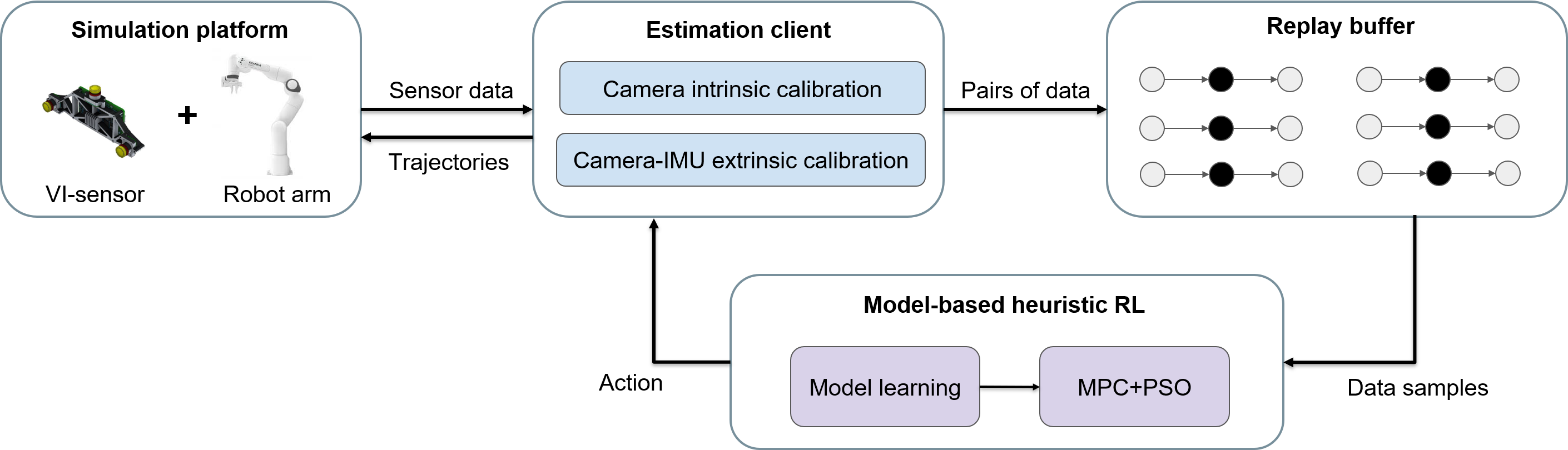}  
  \caption{Overview of our calibration framework. The trajectories (actions) the agent chooses to take and their simulated calibration performance are recorded in real-time. This data is then used to train the agent with model-based \ac{rl}.}
  \label{fig:framework}
\end{figure}

An overview of the proposed learning framework is shown in Figure~\ref{fig:framework}.
Given a set of trajectory parameters, the simulation platform executes the trajectory and records the resulting sensor data.
The estimation client then computes the calibration parameters based on the sensor data and transforms all the results into training data according to the \ac{mdp} formulation of the problem.
Finally, the agent samples from the recorded training data to learn the optimal trajectories using model-based heuristic \ac{rl} and chooses an action to execute. 

\subsection{Visual-Inertial Calibration}
Our estimator follows the Kalibr framework~\citep{furgale2013unified,furgale2012continuous}, where the Levenberg-Marquardt algorithm is applied to minimize the loss between the obtained and predicted measurements to maximize the likelihood of the unknown parameters $Pr(X,\theta|D,L)$.
Here, $X$ is the estimated pose trajectory, $\theta$ depicts the calibration parameters, $D$ are the measurements of a \ac{vi} sensor consisting of images and inertial measurements and $L$ is the known position of landmarks on the calibration target.
As explained in detail in~\citep{schneider2019observability}, the covariance matrix of the known parameters $\Sigma_{X\theta}$ can be obtained from the Jacobian of all error terms and the stacked error covariances.
The covariance of the calibration parameters $\Sigma_\theta$ can be extracted from $\Sigma_{X\theta}$ and further normalized to $\overline{\Sigma}_\theta$.
The information gain can then be evaluated with the following metrics:
\begin{itemize}
    \item \textbf{A-Optimality:} $H_{Aopt} = trace(\overline{\Sigma}_\theta)$
    \item \textbf{D-Optimality:} $H_{Dopt} = \det(\overline{\Sigma}_\theta)$
    \item \textbf{E-Optimality:} $H_{Eopt} = \max(eig(\overline{\Sigma}_\theta))$
\end{itemize}
Minimizing these $H$ metrics leads to the maximization of information gain and furthermore can be used for the reward design of our proposed method.

\subsection{MDP Model for Learning to Calibrate}
\label{subsec:POMDP}

The whole calibration process is modeled as an \ac{mdp}. 
The process description includes a state $S_t$, action $A_t$, transition model $S_{t+1} = f(S_t,A_t)$, and reward $R_t$ at each time step $t$.
We define the action $A_t$ at each time as a looped parameterized trajectory with the same start and terminal pose.
Each action represents a trajectory subsequence, and these subsequences are concatenated to form the final calibration trajectory. At each step, the calibration process needs to be rerun over the entire sequence and cannot be run only over the newly acquired measurements.
The trajectory poses \{$[x_j, y_j, z_j,\alpha_j, \beta_j, \gamma_j]^\top\}_{j=1:J}$ are parameterized by $\{\sum_{q=1,2,4}\bm{a}_q(1-\cos \frac{2q\pi j}{J})+\bm{b}_q\sin \frac{2q\pi j}{J}\}_{j=1:J}$, where $J$ is the number of waypoints inside one action. $\bm{a}_q$ and $\bm{b}_q$ are $6\times1$ vectors.
% a total of
Thus, for one trajectory 36 parameters are needed, which is the action space dimension of the \ac{mdp}.
The reason to choose $\sin$ and $\cos$ as basis functions is that they are capable of representing many good empirical trajectories for calibration.
%

%they are symmetric as well as periodic, and also reduce the unnecessary freedoms. Furthermore, these basis functions are capable of representing most empirical trajectories for calibration.

% \begin{figure}[t]
% \centering
%  \vspace{-0.2cm} 
%  \setlength{\belowcaptionskip}{-0.2cm}
% \begin{minipage}[t]{0.48\textwidth}
% \centering
% \includegraphics[width=0.9\linewidth]{POMDP.png} 
% \caption{\ac{mdp} for calibrarion}
% \label{fig:pomdp}
% \end{minipage}
% \begin{minipage}[t]{0.48\textwidth}
% \centering
% \includegraphics[width=0.83\linewidth]{simulation platform.png}
% % \caption{Simulation platform for visual-inertial system calibration}
% \caption{Simulation platform}
% \label{fig:simulation}
% \end{minipage}
% \end{figure}

%
Let $D_t$ be the measurements acquired with action $A_t$, and $Y_t$ be the vector that stacks all the calibration parameters and their information gain status, then $Y_t = [\theta_t^*, O_t]^T = Cal(\bigcup_{i=0}^{t-1} D_i)$, where $Cal$ represents the calibration process that returns the predicted calibration parameters $\theta^*$ and their respective information gain status $O_t$.
For camera calibration $O_t$ contains the progress of the coverage of the horizontal axis, the vertical axis, size and skew.
For camera-IMU calibration, $O_t$ is composed of the eigenvalues of the covariance matrix for extrinsics, used in the optimization process of the calibration tool.

Ignoring the sensor noise, we assume the action history sequence together with the calibration status determine the measurement sequence: $\bigcup_{i=0}^{t-1} D_i = h(A_{0:t-1}, Y_{0:t-1})$, where $h$ represents an unknown mapping.
In this way, the state $S_t$ is defined as the concatenation of all actions and calibration results of previous time steps: $S_t = [A_{0:t-1}, Y_{0:t}]$.
%, where information gain and calibration status in $Y_{0:t}$ further helps with decision making.
%
The state transition satisfies the Markov property and the transition model becomes
\begin{align}
    S_{t+1} = A_{0:t}\cup Y_{0:t+1} = A_{0:t-1}\cup Y_{0:t}\cup Y_{t+1} \cup A_t = S_t\cup Cal(h(A_{0:t}, Y_{0:t}))\cup A_t
    \label{eq:trans}
\end{align}
where the right hand side of \eqref{eq:trans} only depends on $S_t$ and $A_t$.

% The observation $Y_t$ of the \ac{pomdp} is a vector that stacks all the calibration parameters and their information gain status.
% %
% This is determined purely by the current state and the observation model
%
% \begin{align}
%     Y_t = [\theta_t^*, O_t]^T = Est(\bigcup_{i=0}^{t-1} D_i) = Est(S_t),
% \end{align}

Finally, the reward at each step is composed of four parts: empirical reward $e_t$, information gain for the calibration parameters $o_t$, trajectory length $l_t$ and relative calibration error $d_t$.
The empirical reward encodes intuitive requirements such as image view coverage with target observations.
The information gain includes different evaluation metrics such as the determinant, trace, and eigenvalues of the aforementioned covariance matrix for the extrinsics.
The trajectory length $l$ is computed by summing up the position and Euler angle distances between each two neighboring waypoints
\begin{align}
    l = & \sum_{j=1}^J(\|x_j-x_{j-1},y_j-y_{j-1},z_j-z_{j-1}\|_2+C\|\alpha_j-\alpha_{j-1},\beta_j-\beta_{j-1},\gamma_j-\gamma_{j-1}\|_2),
    \label{eq:path_length}
\end{align}
where $C$ is a tuneable weighting factor to balance the importance between rotation and translation.
The calibration errors are defined as the Euclidean distance between the calibration result $\theta^*$ and ground truth $\theta$.
The relative calibration error is further normalized by the norm of the ground truth.
In a real world scenario, where the ground truth calibration parameters are not available, this reward term can instead be computed using the reprojection error of the calibration process, as it directly measures how close to the ground truth calibration we currently are.
Finally, the reward is the weighted sum of increments of each term at each time step
\begin{align}
    R_t = \eta_1\Delta e_t + \eta_2\Delta o_t - \eta_3\Delta d_t - \eta_4\Delta l_t,
\end{align}
where $\eta_1,\cdots,\eta_4$ are positive weights that can be manually tuned separately.

\subsection{Model-based Heuristic Reinforcement Learning with PSO}

\begin{figure}[t]
  \centering
  \setlength{\belowcaptionskip}{-0.4cm}
  % include first image
  \includegraphics[width=0.75\linewidth]{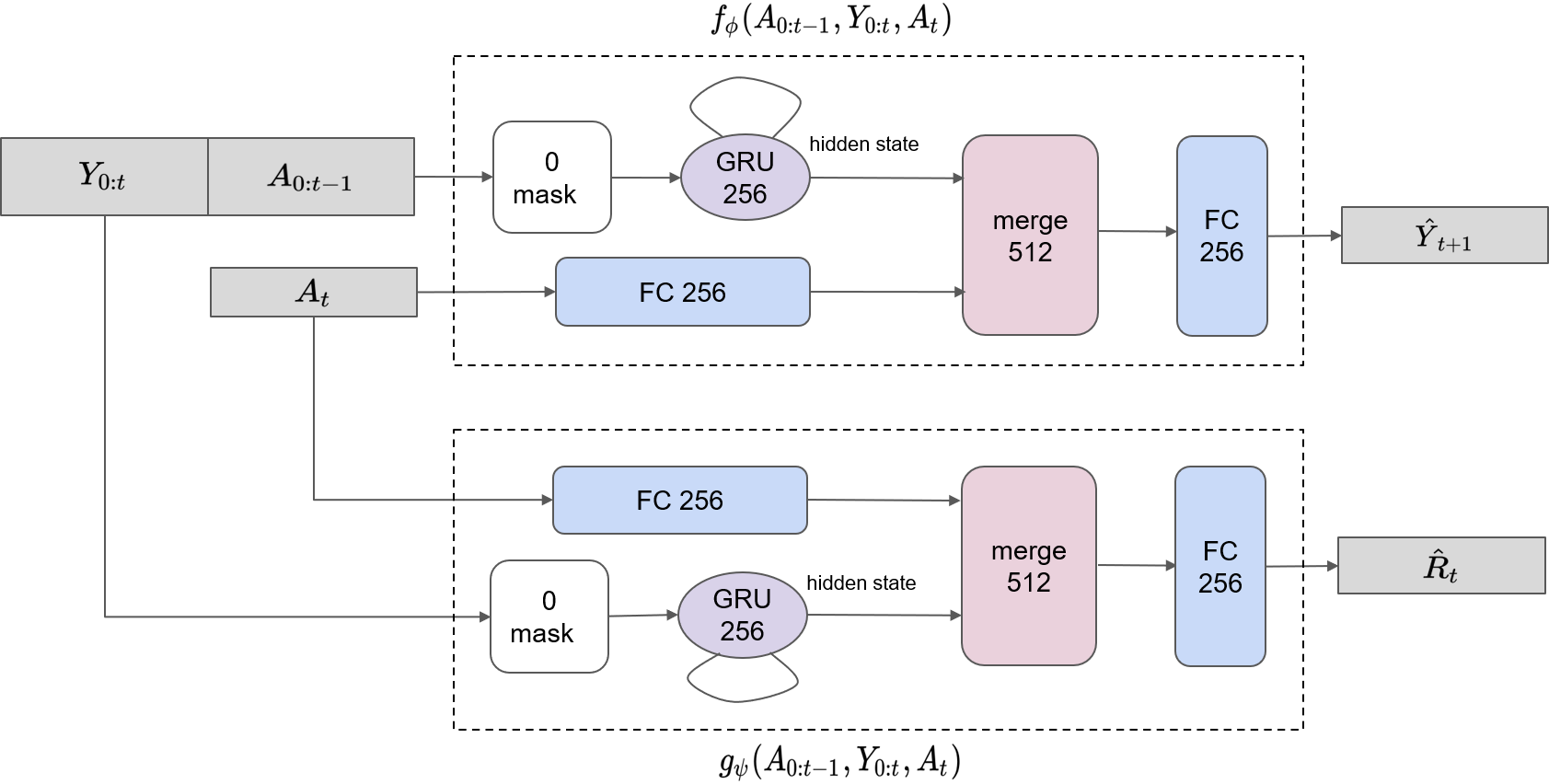}  
  \caption{Network architecture of the dynamics model (top) and reward model (bottom).}
  \label{fig:network}
\end{figure}

\textbf{Neural network dynamics and reward functions}:
We parameterize both the dynamics model and reward model as neural networks.
The network structures are shown in Figure~\ref{fig:network}.
Each model contains a recurrent neural network~\citep{chung2014empirical} to encode a hidden state to compress the information acquired so far, given the action history sequence $A_{0:t-1}$ and the calibration status sequence $Y_{0:t}$ as inputs.
The reward model then feeds the hidden state and current action input $A_t$ to a fully-connected network to predict the current reward $\hat{R}_t = g_\psi(A_{0:t-1},Y_{0:t},A_t)$.
With the same input, the dynamics model predicts the next calibration status $\hat{Y}_{t+1} = f_\phi(A_{0:t-1},Y_{0:t},A_t)$, where $\psi$ and $\phi$ are the weights of the two neural networks.
Given a batch of training data $\{Y_{0:t}^i, A_{0:t-1}^i, A_t^i, Y_{t+1}^i, R_t^i\}_{i=1}^N$, both models are trained using \ac{sgd} to minimizing the mean squared dynamics and reward errors 
\begin{align}
    \epsilon_{dyn} &= \frac{1}{N}\sum_{i=1}^N\|Y_{t+1}^i-\hat{Y}^i_{t+1}\|^2 \label{eq:dyn_f},\\
    \epsilon_{reward} &= \frac{1}{N}\sum_{i=1}^N\|R_{t}^i-\hat{R}^i_{t}\|^2 \label{eq:rew_f}
\end{align}
respectively.
A higher model prediction accuracy is fundamental to the performance of the learned action sequence.

\textbf{Model-based open-loop optimization}:
With the learned reward and dynamics model, we use a \acf{mpc} to control the agent.
In the \ac{mpc} framework, an open-loop action sequence $A^*_{t:T}$ from the current time step $t$ to the end time $T$ is first optimized to maximize the predicted future sum of rewards. 
Then only the first action $A_t$ is executed and the corresponding new states and rewards are obtained. 
The optimal action sequence until the end $A^*_{t+1:T}$ is then recalculated and the next action is executed. 
For our problem, the open-loop optimization to solve the optimal future action sequence $A^*_{t:T}$ at each time step $t$ is formalized as
\begin{align}
    A^*_{t:T} = \argmax_{A_{t:T}} \sum_{\tau=t}^T g_\psi(A_{0:\tau-1},Y_{0:t},\hat{Y}_{t+1:\tau},A_\tau),
    \label{eq:optimization}
\end{align}
where $\hat{Y}_{t+1:\tau}$ is predicted using the dynamics model $f_\phi$ for each time step.

This non-linear optimization problem is solved using a modified version of the \ac{pso} algorithm.  
For a particle swarm with $M$ particles in total, each particle position $P^t_{i\in\{1,2,\cdots,M\}}$ represents a potential solution of actions $A^i_{t:T}$. 
For the position update, the velocities $v$ of particles of the original \ac{pso} algorithm include 3 components: social component, cognitive component, and inertia~\citep{kennedy1995particle}. 
In our method, the cognitive component is modified to be the gradient of the optimization function, rather than the direction towards the local best position the particle has ever visited. 
This is because the gradients are more informative to indicate the local optimal position.
Furthermore, the gradients can be obtained from the learned model as $\mu_i = \nabla_{A_{t:T}} \sum_{\tau=t}^T g_\psi(A_{0:\tau-1},Y_{0:t},\hat{Y}_{t+1:\tau},A_\tau)|_{A_{t:\tau}=P^t_i}$.
In this way, the particles move according to the following rules at each optimization iteration $\iota$
\begin{align}
    \prescript{\iota}{}v_i &= \omega_0\prescript{\iota-1}{}v_i + c_1 (G_{best}-\prescript{\iota-1}{}P^t_i) + c_2 \prescript{\iota-1}{}\mu_i \label{eq:PSO1}\\ 
    \prescript{\iota}{}P^t_i &= \prescript{\iota-1}{}P^t_i + \prescript{\iota}{}v_i,
    \label{eq:PSO2}
\end{align}
where $G_{best}$ represents the global best position all the particles have covered. The first two terms in Equation~\eqref{eq:PSO1} represent the inertia and social components, the last one is the modified cognitive component.
The parameters $\omega_0,c_1, c_2$ are coefficients that can be tuned to trade-off between exploration and exploitation.

\begin{figure}[t]
  \centering
  \setlength{\belowcaptionskip}{-0.5cm}
  % include first image
  \includegraphics[width=0.85\linewidth]{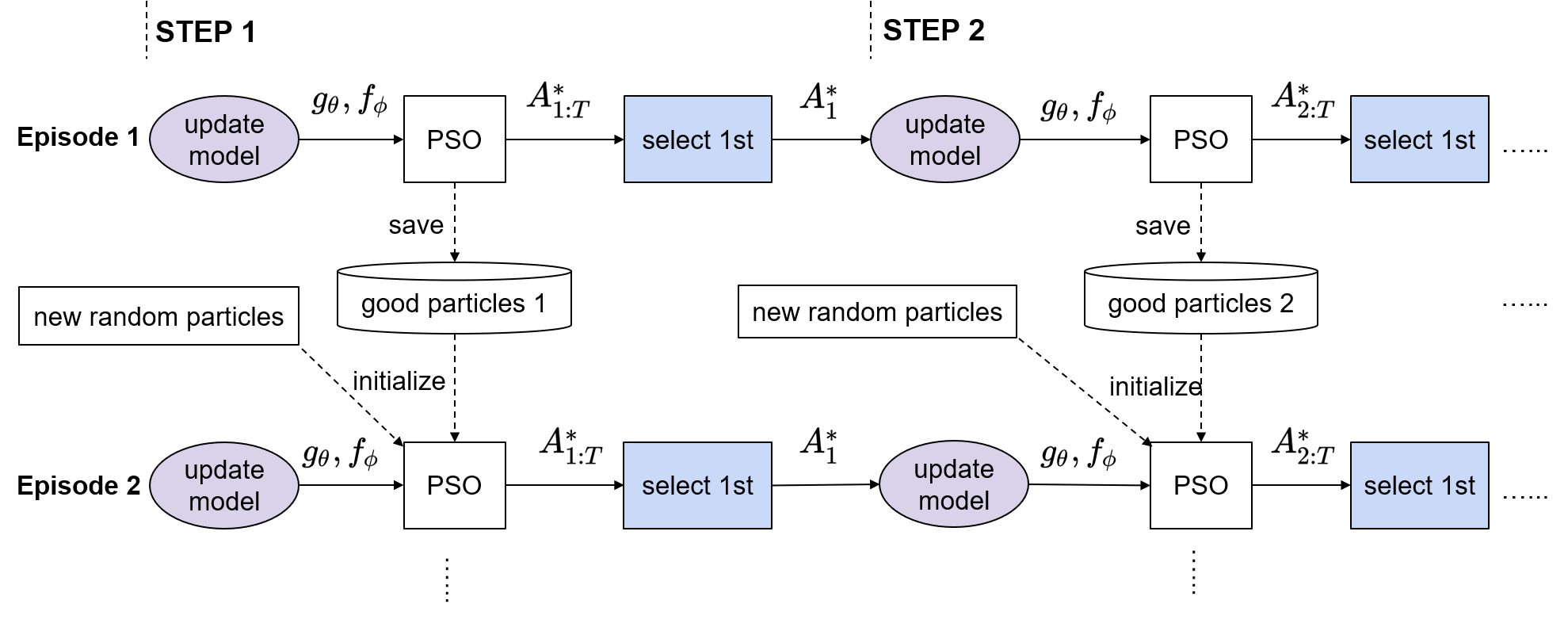}  
  \caption{Particle initialization and action selection in subsequent episodes.}
  \label{fig:particles}
\end{figure}

Due to high computational costs, at each time step, only a small number $I$ of iterations are used for \ac{pso}.
A good solution cannot be guaranteed with a limited number of iterations and random initialization, but the performance can be improved by using previous optimization results to initialize new particles.
Although the optimization problems are different between different time steps in the same episode (the lengths of open-loop action sequences to optimize are different), they are similar between the same time step of different episodes.
Therefore, we save the positions of the top $K$ particles after each time step as $P^t_\text{best}$, for initialization of new particles of the same time step in subsequent episodes.
The other $M-K$ new particles are randomly initialized to enable exploration and avoid local minima.

\textbf{Model predictive control}:
After $I$ iterations and obtaining the final position of all particles $P^t_{1:M}= \{A^i_{t:T}\}_{i=1}^M$ at each time step, a particle $A^*_{t:T}$ is selected as the open-loop control solution.
Different policies are applied to select the particle for training and testing in our method. 
For training episodes, in order to encourage exploration, the action sequence is randomly selected from the top $W$ particles with the the highest predicted reward sum, rather than always selecting the globally best particles. 
To prevent the top $W$ particles from converging to the same point, $W>K$ should be satisfied.
In this way, the $W$ particles contain not only the $K$ particles that have been optimized continuously among episodes but also some new particles that are initialized randomly in the current episode.
For testing, only the particle with the highest predicted sum of rewards is selected.
Following the \ac{mpc} rule, we only extract the first action $A^*_t$ from $A^*_{t:T}$ for execution. 
An overview of the particle initialization and action selection framework is shown in Figure~\ref{fig:particles}.
The entire algorithm is summarized in Algorithm~\ref{algo:MBHRL}.  

\begin{algorithm}
  \small
  \caption{Model-based Heuristic Reinforcement Learning}\label{euclid}
  \begin{algorithmic}[1]
      \State set $T$, $M$, $K$, $W$ and $I$, where $K,W<M$ 
      \State initialize $g_\psi$, $f_\phi$ and randomly initialize $\{P_\text{best}^t\}_{t=0}^{T}$ 
      \State gather dataset $D_{RL}$ with random trajectories
      \For{each episode}
        \State reinitialize the \ac{mdp}
        \For{t=0 \textbf{to} $T$}
            \State train $g_\psi$ and $f_\phi$ by applying SGD to \eqref{eq:dyn_f}-\eqref{eq:rew_f} with $D_{RL}$
            \State update $Y_{0:t}$, $A_{0:t-1}$
            \State initialize $P^t_{1:M}$ by assigning $P^t_{1:M} = \{P^t_\text{best}$, randomly sampling $P^t_{K+1:M}$\}
            \For{$\iota$=0 \textbf{to} $I$}
            \State update $P^t_{1:M}$ using \eqref{eq:PSO1}-\eqref{eq:PSO2}
            \EndFor
          \State sort $P^t_{1:M}$ descendingly by the predicted reward sum
          \State update the best K particles $P^t_\text{best}$ = $P^t_{1:K}$
          \State randomly select $A^*_{t:T}$  from $P^t_{1:W}$
          \State execute $A^*_t$ and obtain $Y_{t+1}$, $R_t$
          \State add $\{Y_{0:t},A_{0:t-1},A_t,Y_{t+1},R_t\}$ to $D_{RL}$
          \EndFor
      \EndFor
      \State \textbf{return} $g_\psi$ and $f_\phi$
  \end{algorithmic}
  \label{algo:MBHRL}
\end{algorithm}
%===============================================================================

% \section{Citations}
% \label{sec:citations}

% 	Citations can be made using either \textbackslash citep\{\} or \textbackslash citet\{\}, depending from the appropriateness. To avoid the citation moving to the next line, it is often a good practice to replace the space before with a tilde (\~{}) character.
% 	Example 1: ``CoRL is the best conference ever, as discussed in~\citep{Calandra2016}.''
% 	Example 2: ``\citet{Calandra2016} proved, both theoretically and numerically, that CoRL is the best conference ever.''

%===============================================================================

\section{Implementation}
\label{sec:implement}

% \subsection{Simulation}
% We evaluate our proposed framework by experiments in the simulation platform. The platform is built based on Gazebo, which includes a visual-inertial system, a checkerboard, and a FRANKA EMIKA Panda robot arm. The visual-inertial system, equipped with a monocular camera and an IMU, is mounted on the end-effector of the robot arm. The settings of camera and IMU sensor in the simulation are shown in the supplementary material. To increase robustness, both the camera intrinsics and camera-IMU extrinsics are sampled from gaussian distributions in every episode. As mentioned in section 3.2, we use sums of sinusoidal and cosinusoidal functions of time for position and orientation with 36 parameters to describe the trajectory. The end-effector, mounted with sensors, moves following the given trajectory in front of the checkerboard, as is shown in Figure \ref{fig:simulation}. 

We evaluate our method by performing experiments in a simulation platform based on Gazebo~\citep{koenig2004design}.
This includes a checkerboard, and a \ac{vi} sensor consisting of a pinhole camera and an \ac{imu} mounted on the end-effector of a FRANKA EMIKA Panda robot arm. 
To increase robustness and generalization, both the camera intrinsics and camera-IMU extrinsics are sampled from Gaussian distributions in every episode.
Each interaction episode is an independent calibration process without sharing measurement with other episodes.
However, the training process is off-policy so that all recorded interaction data is utilized to update the model to improve sample efficiency.  
For the camera intrinsic calibration, we limit the maximum time step size $T$ to be 4 (run at most 4 looped trajectories) in each training episode.
The empirical reward is the coverage of the image view with target observations.
The accuracy reward is computed by dividing the decrease in Euclidean distance from the ground truth by the norm of the ground truth.
The reward and dynamic models for intrinsic calibration are trained in total for 1000 episodes.

For the camera-IMU extrinsic calibration, each training episode contains 3 time steps.
The empirical reward includes the entropy of IMU measurement data in 6 dimensions and number of target observations captured per image.
The model for extrinsic calibration is first trained for 900 episodes.
Subsequently, we fine-tune the agent for another 650 episodes by integrating Kalibr~\citep{furgale2013unified,furgale2012continuous} into the framework.
Two items are added to the reward.
One is the information gain reward, which is the negative A-optimality $H_{Aopt}$ computed based on the covariance matrix.
The other one is the calibration accuracy itself.
For \ac{pso}, we use  $M=15$, $K=5$, $W=5$ and $I=5$. 
Finally, it should be noted that when training on real systems, the ground truths are not available.
In this case calibration error could be replaced by the reprojection error given from the calibration process (see Appendix~\ref{apx:add_exp_results}).

\section{Experiments}
\label{sec:result}
In this section, we conduct two groups of experiments to evaluate the learned trajectories for camera intrinsic calibration and camera-IMU extrinsic calibration.
The learned trajectories are extracted from the result actions using an \ac{mpc} framework on the final reward model and dynamics model.
We compare our learned trajectories with empirically handcrafted trajectories and ones with random parameters as baselines.
%
%The detailed parameters of the trajectories are recorded in the appendix. 
%
The path lengths and calibration errors are computed as stated in Section~\ref{sec:approach}.
In addition, we also verified that a favorable policy can be learned when calibration error is substituted by the reprojection error.
These results are shown in Appendix~\ref{apx:add_exp_results}.

\subsection{Camera Intrinsic Calibration}

\begin{table}[htbp]
\footnotesize
\centering
\vspace{-0.2cm} 
\setlength{\belowcaptionskip}{-0.2cm}
\setlength{\tabcolsep}{1.5mm}{
\begin{tabular}{c|cc|ccc}
& \multicolumn{2}{c|}{\textit{Camera Intrinsic Calibration}} & \multicolumn{3}{c}{\textit{Camera-IMU Extrinsic Calibration}}    \\ \hline
& \textbf{Mean error}  & \textbf{Path length}  & \textbf{Mean error} & \textbf{Path length} & \textbf{Mean A-optimality} \\ \hline
Random trajectory      & $\unit[0.560]{\%}$                       & $\unit[9.627]{m}$                        & $\unit[0.396]{\%}$           & $\unit[7.331]{m}$                & $\unit[4.16\cdot 10^{-08}]{}$            \\
Handcrafted trajectory & $\unit[0.196]{\%}$                       & $\unit[11.116]{m}$                       & $\unit[0.340]{\%}$           & $\unit[3.306]{m}$             & $\unit[1.97\cdot 10^{-07}]{}$            \\
\textbf{Learned trajectory}     & $\mathbf{\unit[0.159]{\%}}$                       & $\unit[11.037]{m}$                       & $\unit[0.265]{\%}$           & $\unit[4.367]{m}$                & $\unit[5.83\cdot 10^{-08}]{}$            \\
\textbf{Fine-tuned trajectory}  & ---                         & ---                          & $\mathbf{\unit[0.214]{\%}}$           & $\unit[3.241]{m}$             & $\unit[9.83\cdot 10^{-08}]{}$            \\ \hline
\end{tabular}}
\vspace{0.2em}
\caption{Comparison of mean relative calibration error, path length, and A-optimality between random, handcrafted, learned, and fine-tuned trajectories. Note that the handcrafted and learned trajectories are different for intrinsic and extrinsic calibration. The means are averaged over all intrinsics and extrinsics settings for evaluation. The path lengths are computed by Equation \ref{eq:path_length} where $\unit[C=1]{m/rad}$.}
\label{tab:average}
\end{table}

\begin{figure}[htbp]
  \centering
  % include first image
%   \vspace{-0cm} 
  \setlength{\belowcaptionskip}{-0.2cm} 
  \includegraphics[width=0.85\linewidth]{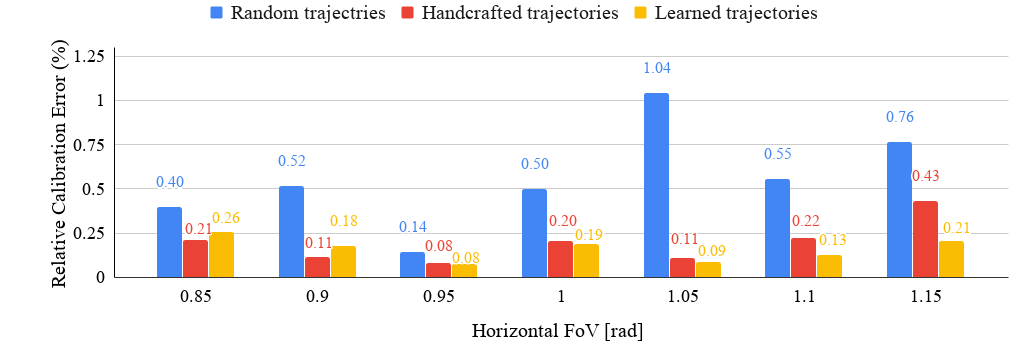}  
  \caption{Relative errors of the intrinsic calibration results w.r.t. the ground truth for cameras with different horizontal \acp{fov}. For each pair of camera setting and policy, we perform the calibration experiments 5 times and calculate the average relative calibration errors w.r.t the ground truth.}
  \label{fig:intrinsic_error}
\end{figure}

We evaluate our policy on cameras with the same image size ($\unit[640]{px}\times\unit[480]{px}$) but different \acp{fov}.
%
%The calibration error is defined as the Euclidean distance from the calibration result to the ground truth. 
%
The relative errors of the intrinsic calibration results w.r.t. the ground truth in different settings of horizontal \acp{fov} are reported in Table~\ref{tab:average} and Figure~\ref{fig:intrinsic_error}.
For smaller \acp{fov}, the handcrafted trajectory renders slightly higher calibration accuracy than learned trajectories.
This may be because, with smaller \acp{fov}, the target appears larger in the image view, which makes it easier for both the learned and handcrafted trajectories to achieve high calibration accuracy.
However, with larger \acp{fov} which make the target smaller in the image view and harder to achieve a high image coverage with target observations, the learned trajectory shows better performance.
%
%The path length are computed by sum of both translation distance (m) and rotation distance (L2-distance in ZYX eular angle space) between neighboring waypoints of the trajectories. 
%
Table~\ref{tab:average} shows that under similar path lengths, our framework could learn how to perform favorable motion trajectories and collect enough measurements efficiently that yield the desired camera intrinsic parameters.

\subsection{Camera-IMU Extrinsic Calibration}

\begin{figure}[htbp]
  \centering
  \vspace{-0.2cm} 
  \setlength{\belowcaptionskip}{-0.2cm}
  % include first image
  \includegraphics[width=0.85\linewidth]{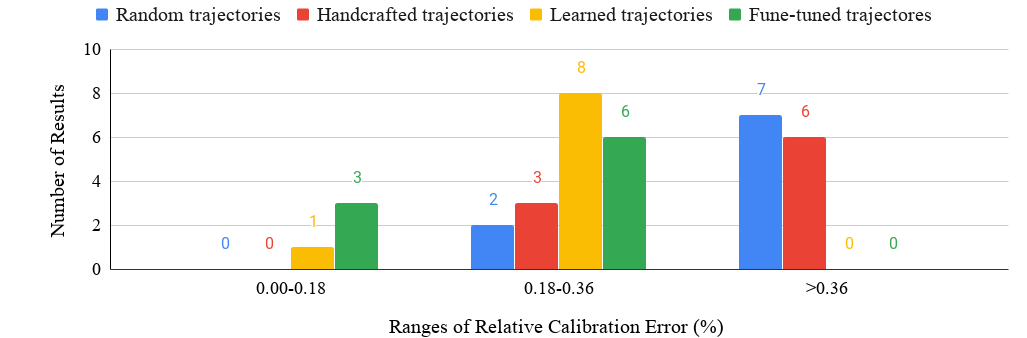}  
  \caption{The distribution of the relative errors of extrinsic calibration results w.r.t. the ground truth. Each trajectory is tested with 9 different pairs of camera intrinsic and \ac{vi} extrinsic configurations.}
  \label{fig:extrinsic_error}
\end{figure}

In this experiment, we train an agent to learn a policy for camera-\ac{imu} extrinsics calibration. As shown in Figure~\ref{fig:extrinsic_error} and Table~\ref{tab:average}, the trajectories generated by our learned policy without fine-tuning can achieve higher calibration accuracies compared with the random and handcrafted trajectories. The low mean A-optimality of learned trajectories can be interpreted as a confidence of Kalibr about the calibration results. 
%
% As mentioned in Section~\ref{sec:implement}, we also fine-tune the trained models for 650 episodes using the extended framework and reward function. 
%
The fine-tuned trajectories perform the best as they achieve the lowest mean error and the shortest path length. The higher mean A-optimality of the fine-tuned trajectory compared with the learned trajectory is possibly caused by the lower path length. A lower path length provides less data, which results in Kalibr being less confident about the calibrations.

%===============================================================================

\section{Conclusion}
\label{sec:conclusion}

In this work, we introduced a novel framework that uses model-based deep \ac{rl} with an adapted version of \ac{pso} for sampling, to generate trajectories for efficiently collecting measurements to calibrate both camera intrinsic and \ac{vi} extrinsic parameters.
Our experiments show that under similar or even shorter path lengths, the trajectories generated by our learned policy can lead to more accurate results and a higher calibration confidence.
%
%With short horizons of our problem, the algorithm will not suffer from vanishing gradients or too large dimensional space for optimization. For problems with longer horizons, investigating combination with model-free \ac{rl} would improve performance.
%
While the proposed model-based \ac{rl} framework is able to achieve state of the art performance in our \ac{mdp} model for \ac{vi} calibration, an interesting avenue for further improvement is to integrate our approach with model-free learners.
%
% When solving problems with a longer horizon, this would help to overcome potential shortcomings, including vanishing gradients and a too large dimensional space for optimization.
%
% Thus, future work involves 
%
% for example, integrating value learning and policy gradients in the learning framework to reduce planning horizons and improve stability.
%
Additionally, model-free \ac{rl} can also be applied for further fine-tuning the learned policies.
%
% At a high level, our method is able to learn how to collect data efficiently for \ac{vi} calibration, which could also extend to other roboic tasks that require certain data to perform optimization. 
%
% In the future, we will explore the high potential of our method in the marker-less self-calibration task.

%===============================================================================

% The maximum paper length is 8 pages excluding references and acknowledgements, and 10 pages including references and acknowledgements

\clearpage
% The acknowledgments are automatically included only in the final version of the paper.
\acknowledgments{This paper was partially supported by ABB Corporate Research, Siemens Mobility GmbH, ETH Mobility Initiative under the project \textit{PROMPT}, and the Luxembourg National Research Fund (FNR) 12571953.}

%===============================================================================

\bibliography{example}  % .bib
\clearpage

\section{Appendix}
\appendix
%===============================================================================

% \begin{abstract}
% % \section{Introduction}
% %
% In this document we describe more details of our work that are not provided in the main paper, including thorough implementation, hyperparameters chosen, and additional experiment results.
% %
% These contents may help interesting readers to reproduce our results, and further support that our simulation is realistic and experiment results are feasible for real systems. The code will be made open-source on publication.
% \end{abstract}

\section{Implementation Details}

\begin{figure}[h]
\centering
\begin{subfigure}[t]{0.285\textwidth}
  \centering
  \includegraphics[width=\linewidth]{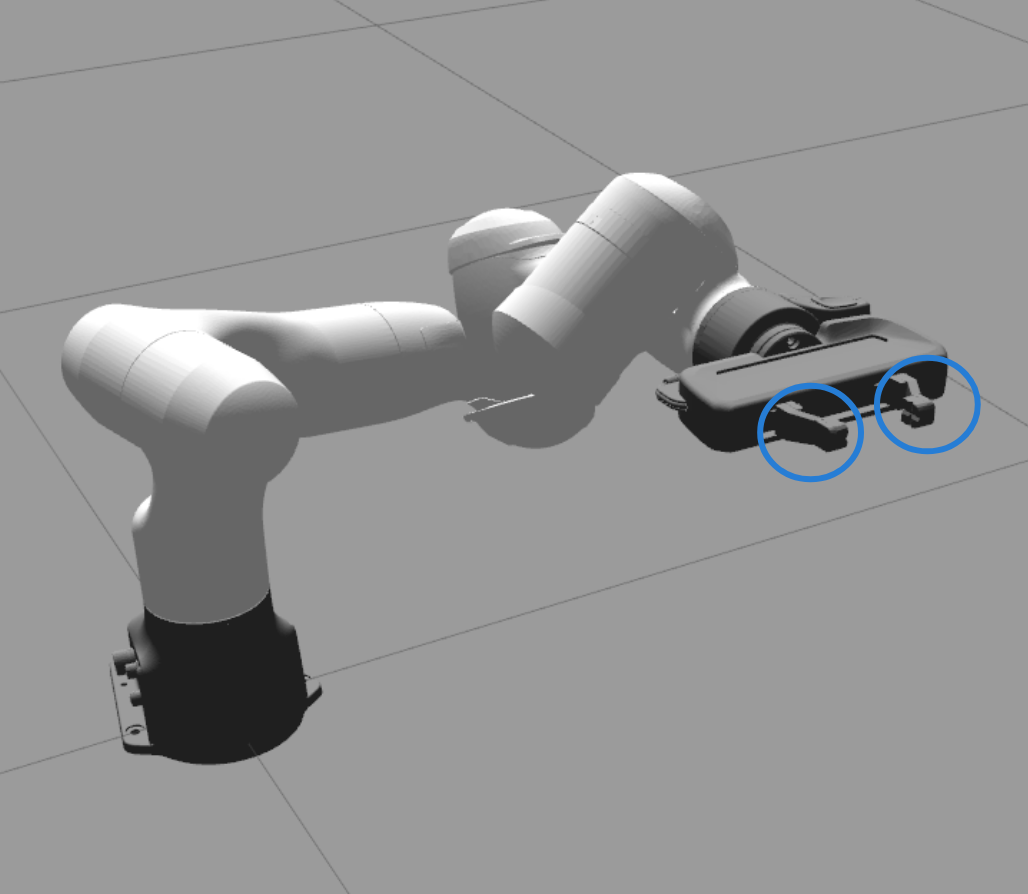}  
  \caption{}
  \label{fig:sima}
\end{subfigure}\hspace{0em}
\begin{subfigure}[t]{0.32\textwidth}
  \centering
  \includegraphics[width=\linewidth]{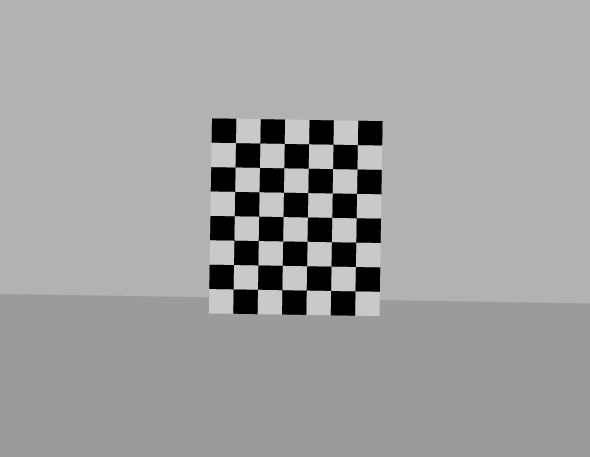}  
  \caption{}
  \label{fig:simb}
\end{subfigure}\hspace{0em}
\begin{subfigure}[t]{0.33\textwidth}
  \centering
  \includegraphics[width=\linewidth]{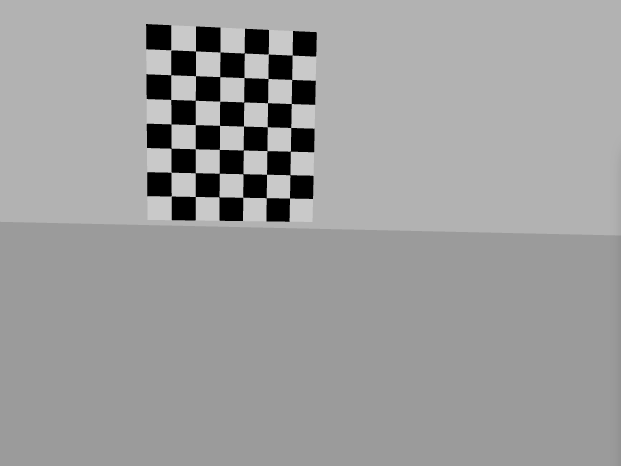}  
  \caption{}
  \label{fig:simc}
\end{subfigure}
\caption{The position of sensors mounted on the end-effector and example image views. (a): positions of the camera (left finger) and IMU (right finger). (b)-(c): example image views from the camera, where (b) is the initial pose, (c) is for a different pose that changes the image view.}
\label{fig:sim}
\end{figure}

\textbf{Simulation settings:}
We use Gazebo~\citep{koenig2004design} for dynamics and sensor simulation. 
Our simulation environment includes a checkerboard, and a VI sensor consisting of a pinhole camera and an IMU mounted on the end-effector of a FRANKA EMIKA Panda robot arm\footnote{https://erdalpekel.de/?p=55}.
%
% For the robot arm, the simulation is set based on \url{https://erdalpekel.de/?p=55}.
%
For the robot arm, the mass of the links is computed according to the total mass of the robot with a uniform density assumption. 
The physical parameters for materials are set by the values for aluminum.
A typical PID controller is set for each joint.
The exact configuration of the nominal VI-sensors is shown in Figure \ref{fig:sima}, which can be modified by changing the camera-IMU extrinsics. 
The detailed sensor settings are shown in Table \ref{tab:IMU} and \ref{tab:camera}. 
We include noise and drift for the IMU and distortion for the camera to achieve more realistic simulation.
During training, the parameters for camera intrinsics and camera-IMU extrinsics are re-sampled from Gaussian distributions after each episode, to ensure that the model learns to generalize well also to other similar sensors.
The parameters for the Gaussian distributions are shown in Table \ref{tab:dist}.
Our target board is a $7\times6$ checkerboard with $\unit[6]{cm}\times\unit[6]{cm}$ squares.
The distance from the target to the robot arm's initial pose is $\unit[2]{m}$.

\begin{table}[!hbt]
\footnotesize
    \centering
    \small
    \begin{tabular}{c c c c c}
    \hline
        update rate & acceleration drift & acceleration noise & augular velocity drift & angular velocity noise\\
        \hline
        200 Hz & $\unit[0.006]{m/s^2}$ & $\unit[0.004]{m/s^2}$ & $\unit[0.000038785]{rad/s}$ & $\unit[0.0003394]{rad/s}$ \\
         \hline
    \end{tabular}
    \caption{Simulation settings for the IMU.}
    \label{tab:IMU}
    %\vspace{-2em}
\end{table}
\begin{table}[!hbt]
\footnotesize
    \centering
    \small
    \begin{tabular}{c c c  c c c c}
    \hline
        update rate & width & height &  norminal horizon FOV & camera model\\
        \hline
        10 Hz & $\unit[640]{px}$ & $\unit[480]{px}$  & \unit[1.0]{rad} & pinhole \\
         \hline
    \end{tabular}
    \caption{Simulation settings for the camera.}
    \label{tab:camera}
    %\vspace{-1em}
\end{table}

\begin{table}[!hbt]
\footnotesize
\centering
\small
\begin{tabular}{c|c|c c c c c c}
\hline
                     & \textbf{Intrinsics}      & \multicolumn{6}{c}{\textbf{Extrinsics}}                                                       \\
\hline
Parameters & Horizontal FOV [rad] & X {[}m{]} & Y {[}m{]} & Z {[}m{]} & Roll {[}rad{]} & Pitch {[}rad{]} & Yaw {[}rad{]} \\
\hline
Mean                 & 1.00           & 0.06      & 0.00      & -0.10     & 0.00           & 0.00            & 1.5708        \\
Std.                  & 0.05           & 0.01      & 0.01      & 0.01      & 0.10           & 0.10            & 0.10         \\
\hline
\end{tabular}
\caption{Gaussian distribution settings for intrinsics and extrinsics parameters during training.}
\label{tab:dist}
%\vspace{-2em}
\end{table}

\textbf{Estimation:}
First, our camera intrinsic calibration is based on the OpenCV~\citep{bradski2000opencv} calibration toolbox. 
The image samples are recorded from the Gazebo camera sensor data. 
Not all samples are included in the database in practice to avoid processing redundant data and unbalancing the calibration. 
Instead, given a new sample, the OpenCV calibration toolbox compares the variety of this sample with all the samples in the database and judges whether the motion speed is sufficiently low between this frame and the previous frame. 
Furthermore, if this sample is different enough from others and the movement speed is not too high, it would be added to the database. 
The algorithm also judges if the samples in the database provide good coverage and variation of target observations to do the calibration by computing how much progress has been made toward adequate variation. 
The coverage of target observations in the image view is computed by the sum of `X', `Y', `size', and 'skew' coverage progress.
When exceeding a certain threshold, the calibration procedure is triggered and the camera parameters are estimated. 

On the other hand, our camera-IMU extrinsic calibration is based on the Kalibr~\citep{furgale2013unified,furgale2012continuous,rehder2016extending} toolbox. 
The data input for calibration are camera and IMU measurements from Gazebo, which are recorded as ROS-bag files.
The information gain is computed based on the covariance matrices extracted from the linear solver of the Kalibr framework. 
During training, the maximum step of optimization for Kalibr is limited to 1 to reduce time cost, while during testing it is set to 10 to achieve accurate calibrations.
\textbf{Trajectory planning and execution:}
We use the Move-It! manipulation software for planning and execution of trajectories of the end-effector.
To enable sequential trajectory execution, at the beginning and end of each loop trajectory (action) at each time step, the end-effector returns to a predefined initial pose. 
The pose is selected to render high controllability so that the end-effector is able to move freely in a larger space around this pose.
At this pose, the target board is also centered in the image view as is shown in Figure~\ref{fig:simb}. 
Given the current action parameters, poses of a sequence of waypoints are first computed based on the expression of the trajectory.
Then the Move-It! planner computes a Cartesian path that follows all those waypoints and executes the trajectory with joint controllers.
To reduce the possibility of getting stuck while executing the trajectories, we limit the maximum absolute value of each element of action to be $0.015$. 
The trajectory parameters for roll, pitch, and yaw angle are multiplied by 2.5, 2.5, and 5 respectively to obtain reasonable scales.

\textbf{Training:}
We use TensorFlow as our deep learning framework.
For camera calibration, we train the reward and dynamics model for 900 episodes.
For the camera-IMU calibration, we first train the reward and dynamics model with only empirical and path length rewards for 1000 episodes.
This training is relatively fast as there is no need for interfacing with the Kalibr toolbox.
Then, the model is fine-tuned for another 650 episodes by including information gain and accuracy reward obtained from Kalibr.
In both cases, we directly extract the action sequence chosen by the MPC controller for the learned model as final resulting trajectories.
All models were trained on a single desktop computer (Intel i7-9750H CPU @ 2.60GHZ) with an NVIDIA
GTX 1660 Ti GPU.

\begin{table}[t]
\centering
\footnotesize
\begin{tabular}{c|cc}
\hline
                                  & Hyperparameter                                         & Value \\ \hline
\multirow{4}{*}{Intrinsic reward} & empirical weight $\eta_1$                                      & 1.0   \\
                                  
                                  & relative error weight  $\eta_3$                                & 1.0   \\
                                  & path length weight $\eta_4$                                    & 0.2   \\
                                  & accuracy bonus                                         & 5.0   \\                             & reprojection error weight $\eta'_3$                                          & 2.0   \\\hline
\multirow{4}{*}{Extrinsic reward} & empirical weight $\eta_1$                                      & 1.0   \\
                                  & information gain weight  $\eta_2$                              & $1\times 10^{8}$   \\
                                  & relative error weight $\eta_3$                                 & 1.0   \\
                                  & path length weight  $\eta_4$                                   & 1.0   \\
                                   & reprojection error weight $\eta'_3$                                          & 2.0\\ \hline
\multirow{2}{*}{Network training} & reward model learning rate                             & $1\times10^{-4}$  \\
                                  & dynamics model learning rate                           & $1\times10^{-4}$  \\ \hline
\multirow{7}{*}{PSO settings}     & social component weight  $c_1$                              & $1\times10^{-5}$  \\
                                  & cognitive component weight  $c_2$                           & $1\times10^{-4}$  \\
                                  & inertia weight  $\omega_0$                                       & $1\times10^{-5}$  \\
                                  & max iterations  $I$                                       & 5     \\
                                  & number of particles $M$                                  & 15    \\
                                  & number of top particles for initialization  $K$           & 5     \\
                                  & number of top particles to select for action execution $W$ & 5     \\ \hline
\end{tabular}
\caption{Hyperparameter settings for reward design, network training and PSO.}
\label{tab:hyper}
%\vspace{-3em}
\end{table}

\section{Hyperparameters}

% \TODO{TODO: POMDP reward weights (1+3+4)}

% \TODO{TODO: Neural network hyperparameters (learning rate 2)}

% \TODO{TODO: PSO hyperparameters (3+4)}

% \usepackage{multirow}
As is shown in Table \ref{tab:hyper}, our reward design for the intrinsic and extrinsic calibration is slightly different. 
For the intrinsic camera calibration, we give an extra bonus for accurate calibration if the relative error is less than $\unit[1]{\%}$. 
For the extrinsic camera-IMU calibration, as we limit the maximum optimization step of Kalibr, obtaining accurate results is not realistic.
Therefore, the reward for accuracy only includes the calibration error.
The weights are chosen to make the scales of different parts of the reward comparable.
We use the same training setup and PSO hyperparameters for both intrinsic and extrinsic calibration. 
For the PSO settings, we set a low weight for the social component to avoid too early convergence. 
In this way, the particles will focus on exploiting their local regions, which benefits searching for the global optimum.

\section{Additional Experimental Results}
\label{apx:add_exp_results}
\begin{figure}[h]
\centering
\begin{subfigure}[t]{0.32\textwidth}
  \centering
  \includegraphics[width=\linewidth]{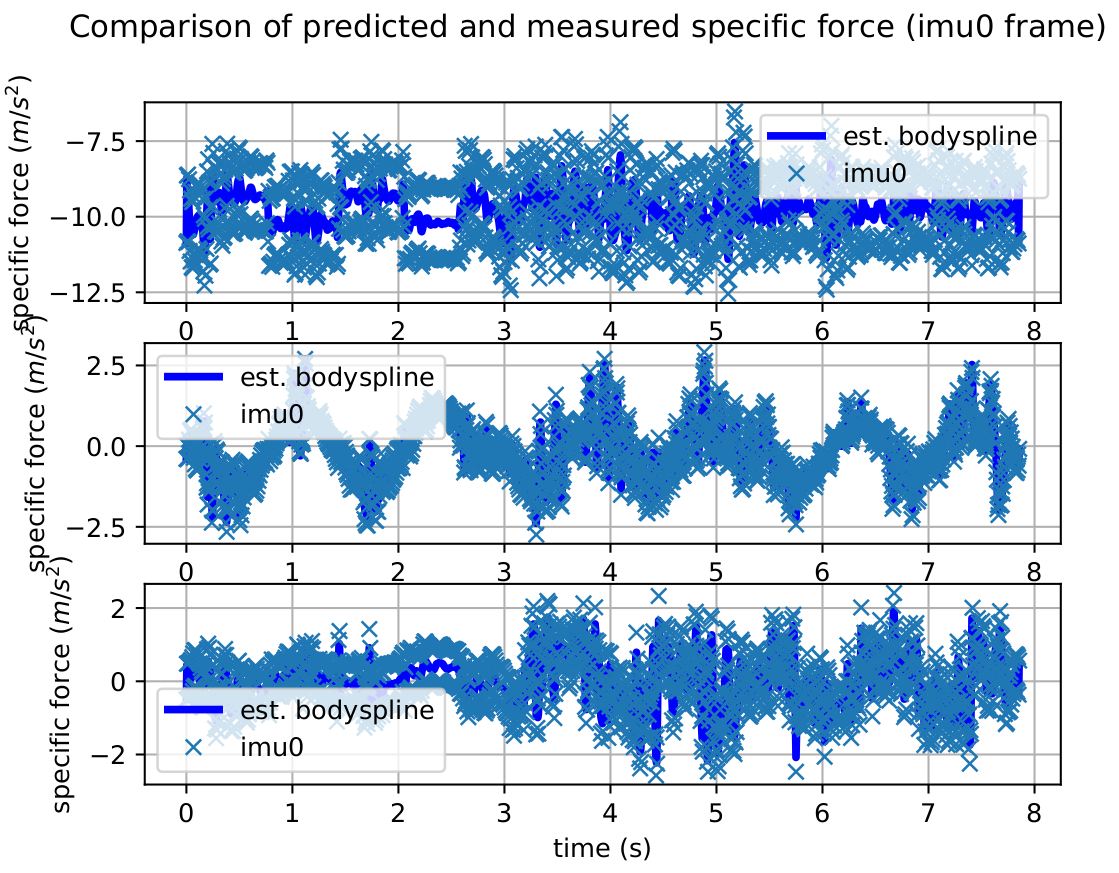}  
  \caption{}
  \label{fig:3a}
\end{subfigure}\hspace{0em}
\begin{subfigure}[t]{0.32\textwidth}
  \centering
  \includegraphics[width=\linewidth]{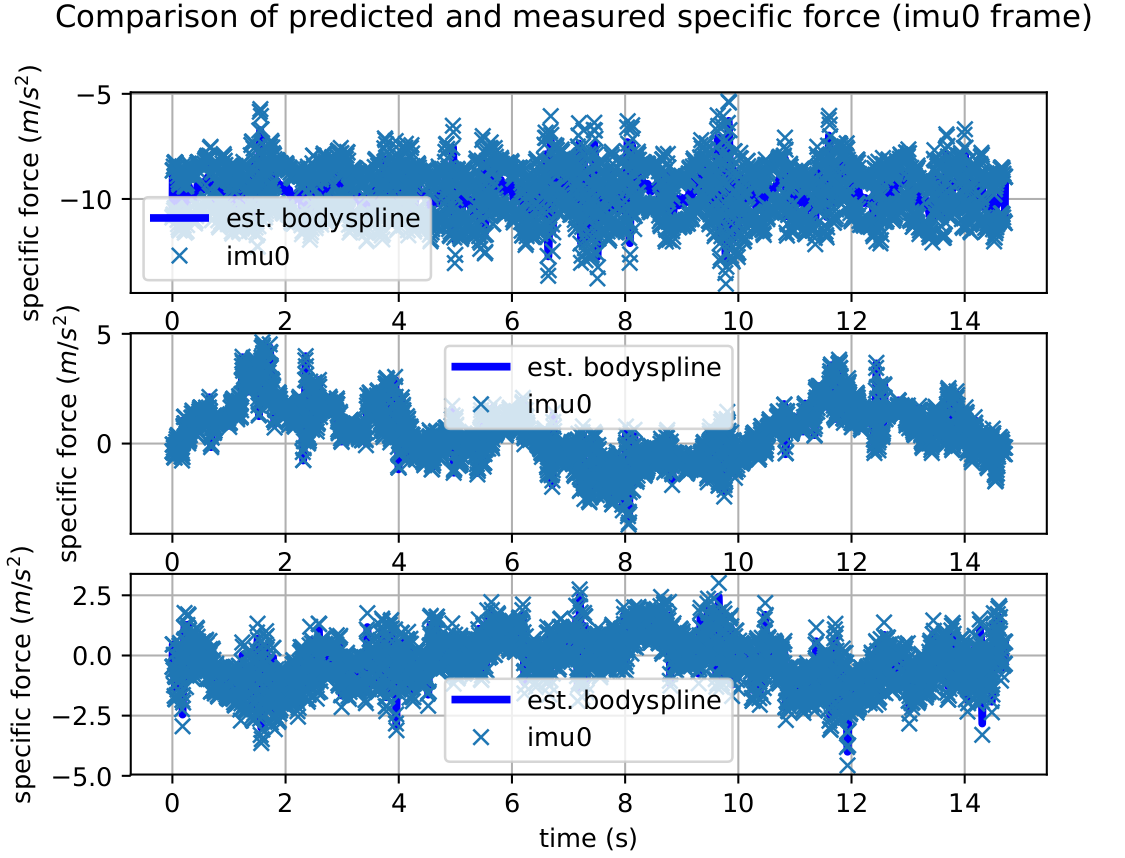}  
  \caption{}
  \label{fig:3b}
\end{subfigure}\hspace{0em}
\begin{subfigure}[t]{0.32\textwidth}
  \centering
  \includegraphics[width=\linewidth]{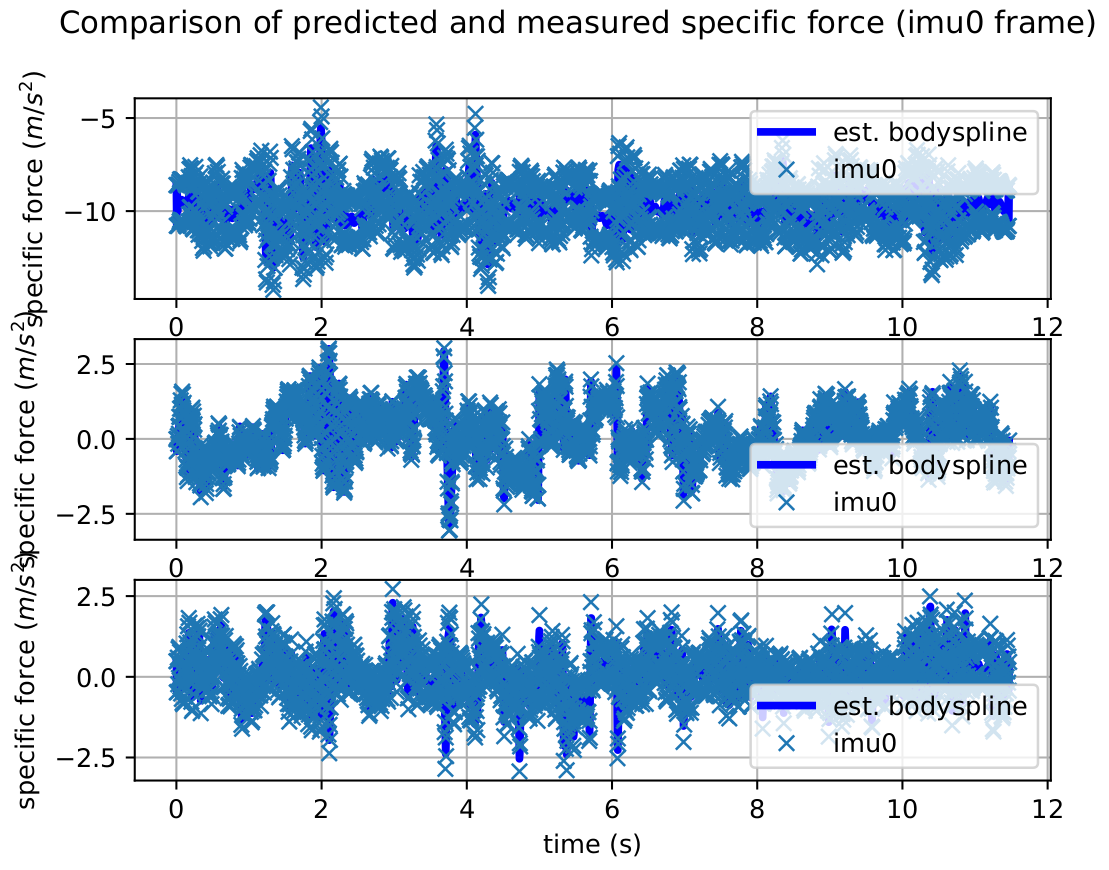}  
  \caption{}
  \label{fig:3c}
\end{subfigure}

\begin{subfigure}[t]{0.32\textwidth}
  \centering
  \includegraphics[width=\linewidth]{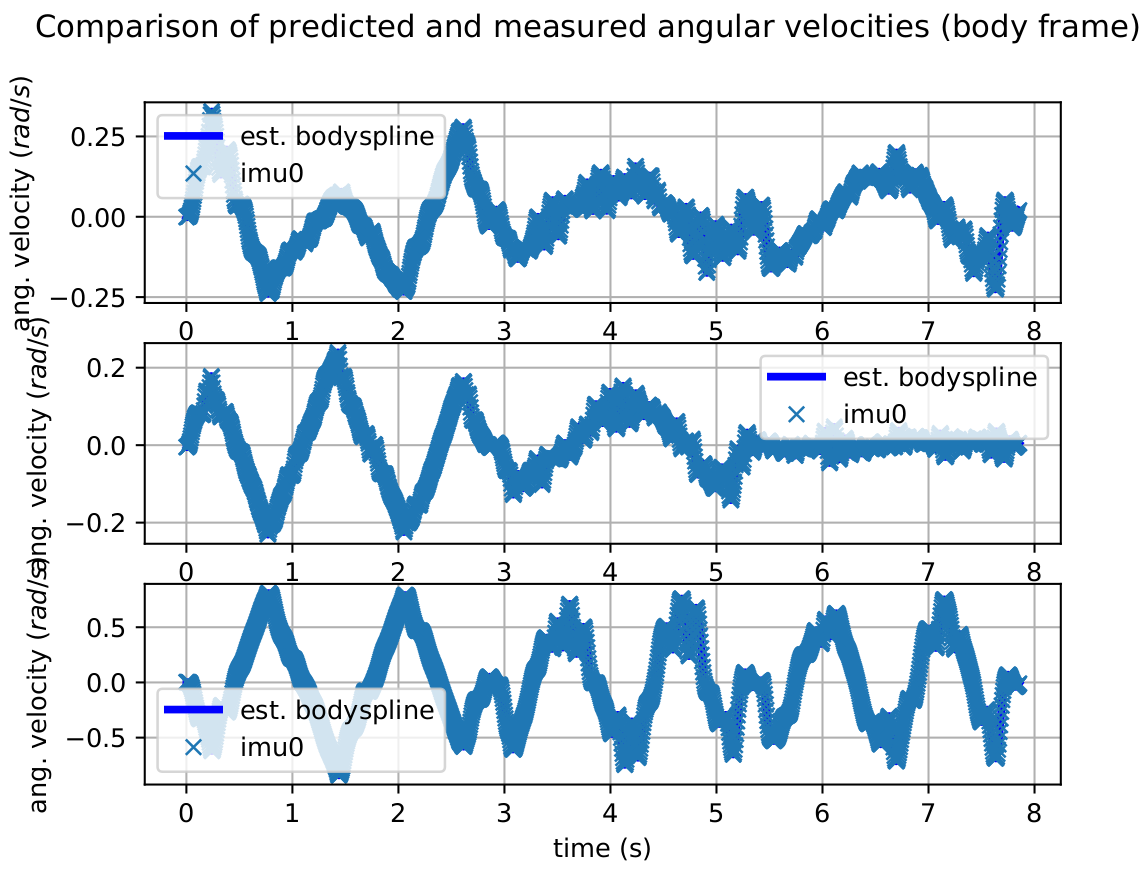}  
  \caption{}
  \label{fig:3d}
\end{subfigure}\hspace{0em}
\begin{subfigure}[t]{0.32\textwidth}
  \centering
  \includegraphics[width=\linewidth]{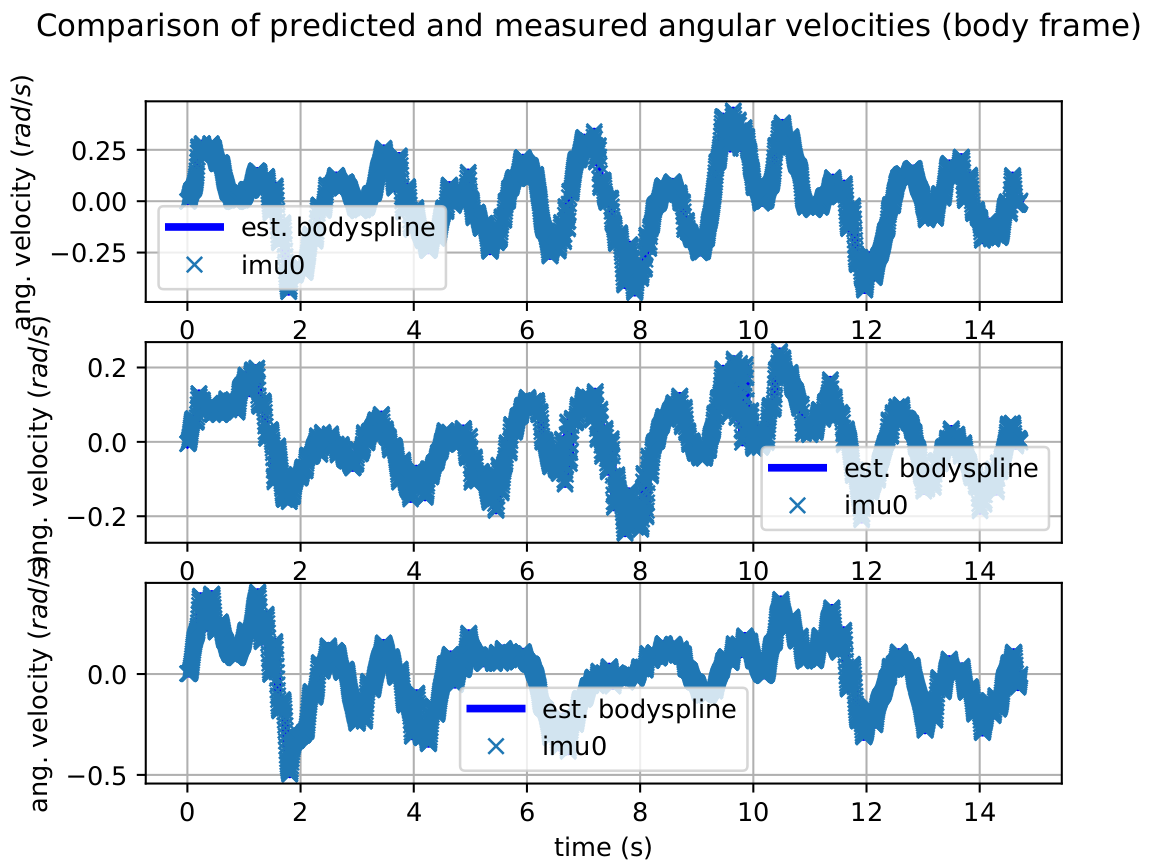}  
  \caption{}
  \label{fig:3e}
\end{subfigure}\hspace{0em}
\begin{subfigure}[t]{0.32\textwidth}
  \centering
  \includegraphics[width=\linewidth]{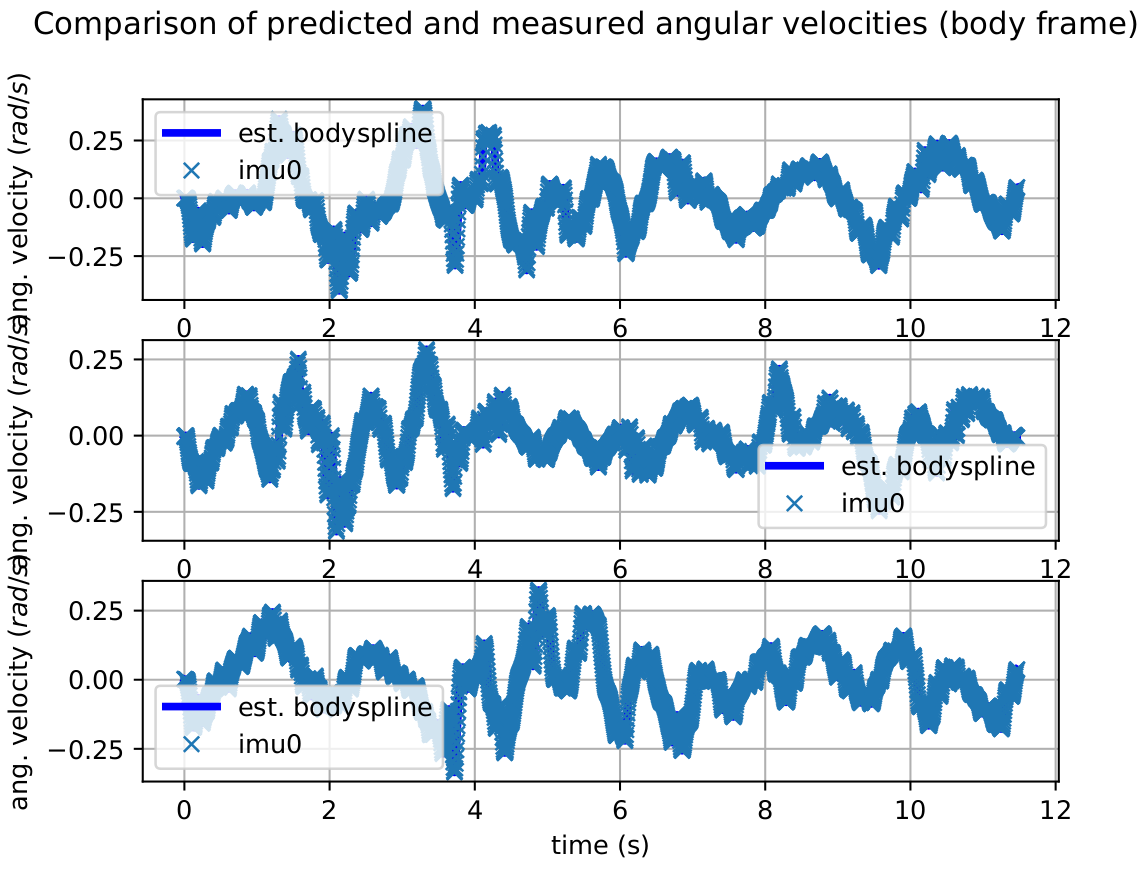}  
  \caption{}
  \label{fig:3f}
\end{subfigure}

\begin{subfigure}[t]{0.32\textwidth}
  \centering
  \includegraphics[width=\linewidth]{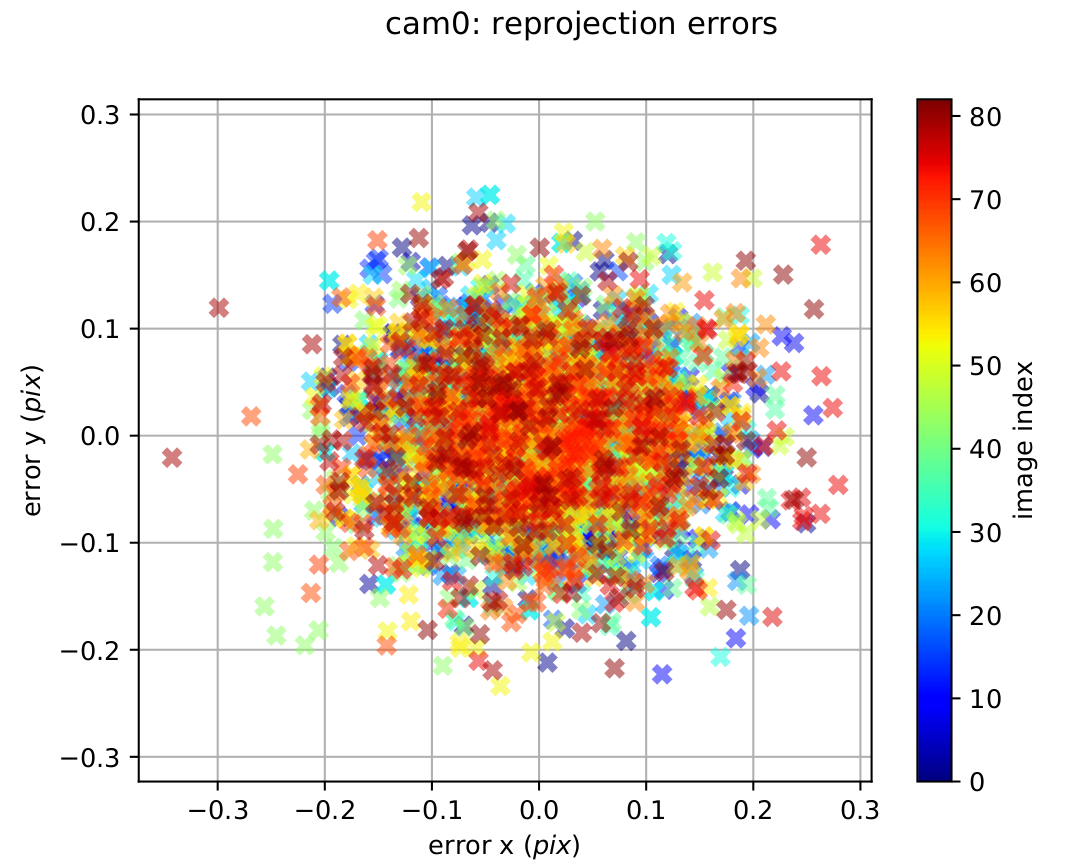}  
  \caption{}
  \label{fig:3g}
\end{subfigure}\hspace{0em}
\begin{subfigure}[t]{0.32\textwidth}
  \centering
  \includegraphics[width=\linewidth]{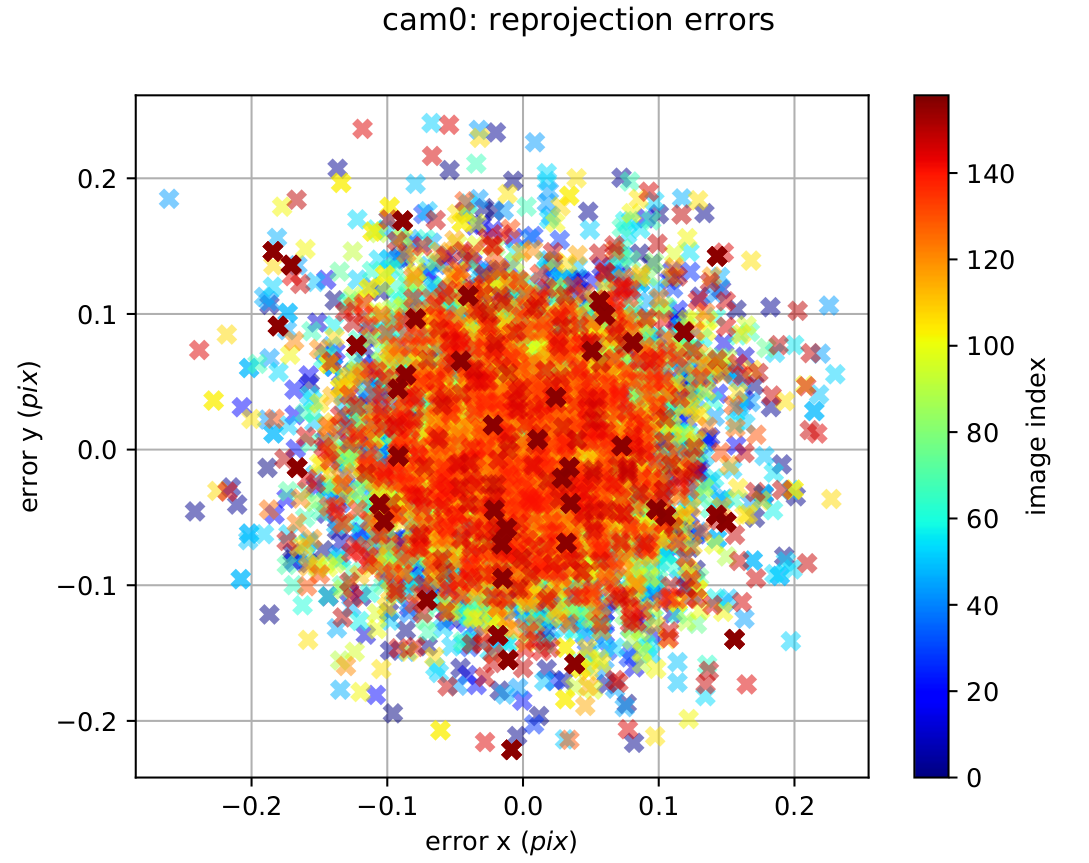}  
  \caption{}
  \label{fig:3h}
\end{subfigure}\hspace{0em}
\begin{subfigure}[t]{0.32\textwidth}
  \centering
  \includegraphics[width=\linewidth]{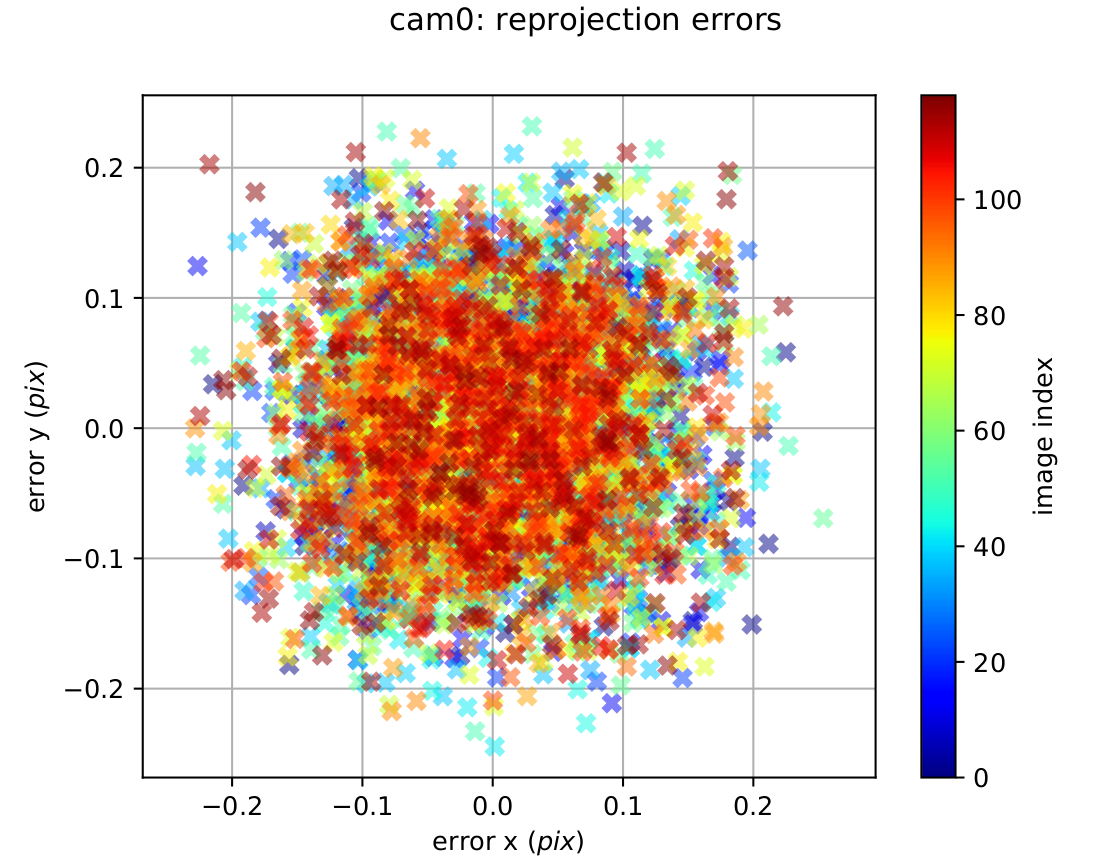}  
  \caption{}
  \label{fig:3i}
\end{subfigure}
\caption{Extrinsic calibration results from the Kalibr toolbox. (a), (d), (g): Results for handcrafted trajectories. (b), (e), (h): Results for learned trajectories without fine-tuning. (c), (f), (i): Results for fine-tuned trajectories.}
\label{fig:3}
\end{figure}

The result outputs by the Kalibr toolbox for the handcrafted, initial learned, and fine-tuned trajectories are shown in Figure \ref{fig:3}. 
High noise and drifts are imposed on the simulated IMU sensor, as can be shown in Figure~\ref{fig:3a}-\ref{fig:3c}, which causes deviation of the measurements from the predicted trajectories.
%In general, all trajectories render good calibration results. 
%However, as can be seen from Figure~\ref{fig:3a}-\ref{fig:3f}, the motions of the learned and fine-tuned trajectories are more exciting within the same ranges.
%
Regarding re-projection errors shown in Figure~\ref{fig:3g}-\ref{fig:3i}, the fine-tuned trajectories achieve smaller and more symmetric error distribution compared to both hand-crafted and fine-tuned trajectories.
Additionally, we extend the result in the main paper by conducting experiments on using reprojection error instead of calibration error during training, as would be done in a real world setup.
The result is shown in the table \ref{tab:ext_average}.
For the camera intrinsic calibration, we achieved similar results. For camera-IMU extrinsic calibration, the trajectories using reprojection error outperform the handcrafted trajectories while being only slightly worse than the trajectories using the calibration error.

\begin{table}[t]
\footnotesize
\centering
\vspace{-0.2cm} 
\setlength{\belowcaptionskip}{-0.2cm}
\setlength{\tabcolsep}{1.5mm}{
\begin{tabular}{c|cc|ccc}
& \multicolumn{2}{c|}{\textit{Camera Intrinsic Calibration}} & \multicolumn{3}{c}{\textit{Camera-IMU Extrinsic Calibration}}    \\ \hline
& \textbf{Mean error}  & \textbf{Path length}  & \textbf{Mean error} & \textbf{Path length} & \textbf{Mean A-optimality} \\ \hline
Random trajectory      & $\unit[0.560]{\%}$                       & $\unit[9.627]{m}$                        & $\unit[0.396]{\%}$           & $\unit[7.331]{m}$                & $\unit[4.16\cdot 10^{-08}]{}$            \\
Handcrafted trajectory & $\unit[0.196]{\%}$                       & $\unit[11.116]{m}$                       & $\unit[0.340]{\%}$           & $\unit[3.306]{m}$             & $\unit[1.97\cdot 10^{-07}]{}$            \\
\textbf{Learned trajectory}     & $\unit[0.159]{\%}$                       & $\unit[11.037]{m}$                       & $\unit[0.265]{\%}$           & $\unit[4.367]{m}$                & $\unit[5.83\cdot 10^{-08}]{}$            \\
\textbf{Fine-tuned trajectory}  & ---                         & ---                          & $\mathbf{\unit[0.214]{\%}}$           & $\unit[3.241]{m}$             & $\unit[9.83\cdot 10^{-08}]{}$            \\ 
\textbf{Learned with reproj error}  & $\mathbf{\unit[0.155]{\%}}$                         & $\unit[9.319]{m}$                          & $\unit[0.268]{\%}$           & $\unit[4.333]{m}$             & $\unit[8.39\cdot 10^{-08}]{}$            \\ 

\hline
\end{tabular}}
\vspace{0.2em}
\caption{Comparison of mean relative calibration error, path length, and A-optimality between random, handcrafted, learned, and fine-tuned trajectories when calibration error is substituted by reprojection error for training.}
\label{tab:ext_average}
\end{table}

%===============================================================================

% The detailed learned and handcrafted action parameters are shown from Table~\ref{tab:int_hand} to Table~\ref{tab:ext_fine}.
% %
% As is mentioned in the main paper, the trajectory poses $\{[x_j, y_j, z_j,\alpha_j, \beta_j, \gamma_j]^\top\}_{j=1:J}$ are parameterized by $\{\sum_{q=1,2,4}\bm{a}_q(1-\cos \frac{2q\pi j}{J})+\bm{b}_q\sin \frac{2q\pi j}{J}\}_{j=1:J}$, where $J$ is the number of waypoints inside one trajectory. 
% %
% For handcrafted sequences, in each trajectory we only choose certain elements of parameters to be non-zero, making trajectories at different time steps cover exciting motions in different axes.
% %
% The scales of parameters for handcrafted trajectories are selected to make their path length comparable to learned and fine-tuned trajectories.
%

\clearpage
% \bibliography{example}  % .bib

\end{document}